\documentclass[journal,twoside,web]{ieeecolor2}
\usepackage{generic}
\usepackage{cite}
\usepackage{amsmath,amssymb,amsfonts}
\usepackage{algorithmic}
\usepackage{graphicx}
\usepackage{makecell}
\usepackage{textcomp}
\usepackage{adjustbox}
\usepackage{multirow}
\usepackage{array}
\usepackage{subfigure}
\def\BibTeX{{\rm B\kern-.05em{\sc i\kern-.025em b}\kern-.08em
    T\kern-.1667em\lower.7ex\hbox{E}\kern-.125emX}}
\begin{document}
\title{CmFNet: Cross-modal Fusion Network for Weakly-supervised Segmentation of Medical Images}
\author{Dongdong Meng, Sheng Li, Hao Wu, Suqing Tian, Wenjun Ma, Guoping Wang, and Xueqing Yan
\thanks{This work was supported by the National Key
Research and Development Program of China
(No.2019YFF01014404) and NSFC (No.62172013).(Corresponding Authors: Sheng Li and Xueqing Yan.)}
\thanks{D. Meng, W. Ma and X. Yan are with the School of Physics, Peking University, Beijing 100871, China (e-mail: dongdongmeng@pku.edu.cn; wenjun.ma@pku.edu.cn; x.yan@pku.edu.cn).}
\thanks{S. Li and G. Wang are with the School of Computer Science, Peking University, Beijing 100871, China (e-mail: lisheng@pku.edu.cn; wgp@pku.edu.cn).}
\thanks{H. Wu is with the Department of Radiotherapy, Peking University Cancer Hospital, Beijing 100142, China (e-mail: hao.wu@bjcancer.org).}
\thanks{S. Tian is with the Department of Radiation Oncology, Peking University Third Hospital, Beijing 100191, China (e-mail: Suqing.tian@bjmu.edu.cn).}
}

\maketitle

\begin{abstract}
Accurate automatic medical image segmentation relies on high-quality, dense annotations, which are costly and time-consuming. Weakly supervised learning provides a more efficient alternative by leveraging sparse and coarse annotations instead of dense, precise ones. However, segmentation performance degradation and overfitting caused by sparse annotations remain key challenges.
To address these issues, we propose CmFNet, a novel 3D weakly supervised cross-modal medical image segmentation approach. CmFNet consists of three main components: a modality-specific feature learning network, a cross-modal feature learning network, and a hybrid-supervised learning strategy.
Specifically, the modality-specific feature learning network and the cross-modal feature learning network effectively integrate complementary information from multi-modal images, enhancing shared features across modalities to improve segmentation performance. 
Additionally, the hybrid-supervised learning strategy guides segmentation through scribble supervision, intra-modal regularization, and inter-modal consistency, modeling spatial and contextual relationships while promoting feature alignment.
Our approach effectively mitigates overfitting, delivering robust segmentation results. It excels in segmenting both challenging small tumor regions and common anatomical structures. Extensive experiments on a clinical cross-modal nasopharyngeal carcinoma (NPC) dataset (including CT and MR imaging) and the publicly available CT Whole Abdominal Organ dataset (WORD) show that our approach outperforms state-of-the-art weakly supervised methods. In addition, our approach also outperforms fully supervised methods when full annotation is used.
Our approach can facilitate clinical therapy and benefit various specialists, including physicists, radiologists, pathologists, and oncologists.
\end{abstract}

\begin{IEEEkeywords}
Modality-specific learning, cross-modal learning, weakly-supervised learning.
\end{IEEEkeywords}

\section{Introduction}
\begin{figure}[!t]
  \centering
    \subfigure
    {
        \includegraphics[width=\columnwidth]{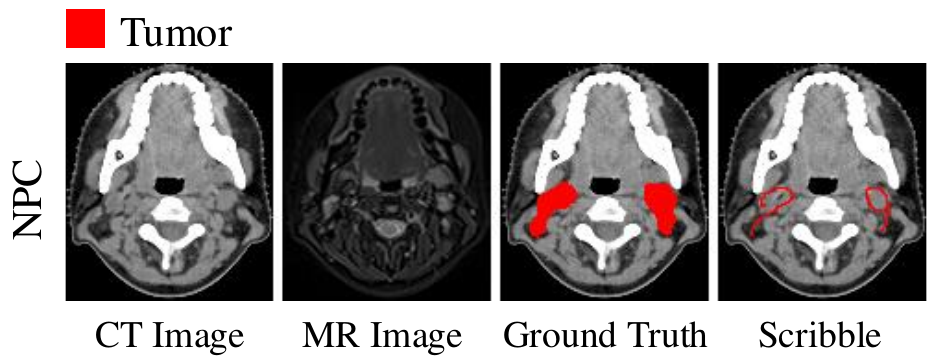}
    }
    \subfigure
    {
         \includegraphics[width=\columnwidth]{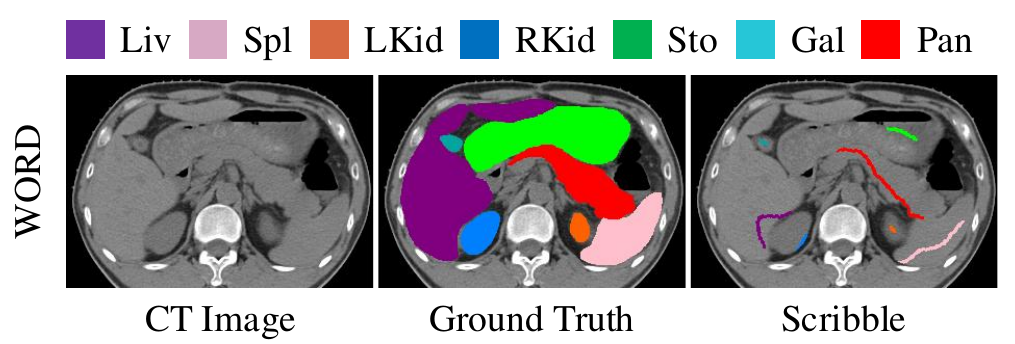}
    }
\caption{Examples of scribble annotations on NPC and WORD datasets. The abbreviations Liv, Spl, Lkid, RKid, Sto, Gal, and Pan are used to denote the liver, spleen, left kidney, right kidney, stomach, gallbladder, and pancreas, respectively. }
\vspace{-0.2in}
\label{fig-samples}
\end{figure}

\IEEEPARstart{f}{ully-supervised}
deep learning methods are generally limited by the time-consuming and costly process of manual annotations \cite{luo202miccai,surveysegmentation2021image}. To address this, many techniques have been developed to reduce the reliance on dense annotations while maintaining high-quality segmentation.
Wherein, weakly-supervised learning leverages weak or sparse annotations, such as image-level annotations \cite{zhou2016imagelevel, Dolz2022weakly}, scribble annotations \cite{han2024dmsps, lee2020scribble2label}, bounding boxes \cite{rajchl2016deepcut} and point annotations \cite{PASEG}. It reduces the annotation effort and has been successfully applied to many segmentation tasks. Among these methods, scribble annotations are particularly flexible and versatile, able to effectively handle complex objects in images \cite{li2024scribformer}. However, scribble-based annotations for 3D object segmentation present two major challenges. Firstly, sketchy scribble annotations make it difficult to accurately capture the detailed variations of the target object, especially in areas with long-range changes, low contrast, and ambiguous boundaries (see Fig. \ref{fig-samples}), often resulting in reduced accuracy. Secondly, with only a limited number of annotated pixels, the segmentation model increases the risk of overfitting, thereby impairing their generalization capabilities. When handling nasopharyngeal carcinoma (NPC) tumor images, irregular boundaries, as well as varying shapes and sizes of tumors, further complicate the segmentation task.

To address both overfitting and the resultant reduced accuracy caused by limited annotated pixels, several strategies have been explored, including pseudo labels \cite{lin2016scribblesup, luo202miccai}, regularization loss \cite{PASEG, EMLoss2004semi, TVjavanmardi2016}, and consistency learning strategies \cite{ustm2022PR, lee2020scribble2label}. 
Pseudo-labeling assigns semantic labels to unannotated pixels, offering advantages for objects with high contrast \cite{lin2016scribblesup, luo202miccai}, yet it can suffer from error propagation and noise for low-contrast regions. Regularization losses curb overfitting by penalizing unannotated pixels \cite{PASEG, EMLoss2004semi, TVjavanmardi2016}, but their efficacy depends on data alignment with assumed conditions. Consistency loss maintains prediction stability under minor perturbations \cite{ustm2022PR, lee2020scribble2label}, although its success hinges on the chosen perturbation strategy and can be challenging for cross-modal data. Despite these limitations, these approaches can somewhat reduce the overfitting of sparse annotations and improve the accuracy and generalization in segmentation.

Cross-modal fusion can enhance weakly supervised segmentation by incorporating detailed information and improving boundary and structure learning. While effective, these approaches still face several challenges, as discussed below.
Most cross-modal fusion methods merge images at the input level to learn shared representations \cite{arrayzhou2019review, Renarticle}, potentially overlooking modality-specific features and resulting in low performance. Hierarchical-level fusion trains separate networks and integrates features layer-by-layer \cite{dolz2018hyperdense,jbhiyang2023mmfusion} but demands complex architectures, increasing computational complexity and the risk of overfitting. Decision-level fusion uses parallel networks and merges outputs \cite{arrayzhou2019review, MAMLzhang2021modality}, risking incomplete use of complementary modality, and error accumulation. Therefore, effectively capturing modality-specific features while leveraging shared cross-modal information is our pathway to achieving high-accuracy segmentation.

Considering scribble-based annotation is convenient and rich in structural information, it is more suitable for non-convex shapes than other weak annotations \cite{han2024dmsps, ustm2022PR}. 
In this paper, we propose a weakly supervised 3D cross-modal segmentation approach for medical images based on only scribble annotations. Our method achieves high accuracy while mitigating overfitting caused by sparse annotations. We specifically validate our approach for NPC segmentation, a widely recognized difficult task, achieving high accuracy comparable to fully supervised state-of-the-art (SOTA) methods. 
To achieve this, we model the characteristics of each modality and design two modality-specific feature learning branches to extract hierarchical modality-specific features from different modalities. Additionally, we introduce a cross-modal feature learning branch to effectively utilize shared information across modalities. 
Furthermore, we propose a hybrid-supervised learning strategy that, besides using scribble supervision, ensures the consistency and stability of cross-branch segmentation results through intra-modal regularization and inter-modal consistency, while also preventing overfitting.

Overall, our main contributions can be summarized as:

\begin{itemize}
   \item[$\bullet$] We present a weakly supervised segmentation technique using scribble supervision for cross-modal medical images that delivers precise and resource-efficient clinical automatic delineation.
\end{itemize}

\begin{itemize}
   \item[$\bullet$] We introduce two modality-specific feature learning branches to model each modality and a cross-modal feature learning branch to integrate shared and complementary information, enhancing segmentation performance. 
\end{itemize}

\begin{itemize}
   \item[$\bullet$] 
    We propose a hybrid-supervised learning strategy combining scribble supervision, intra-modal regularization, and inter-modal consistency constraints to guide segmentation, reduce overfitting, and ensure stable segmentation.
\end{itemize}

\begin{itemize}
   \item[$\bullet$] 
   Our method effectively segments both NPC and general anatomical structures. Validated on a cross-modal clinical NPC  dataset (including CT and MR) and the public WORD dataset, it outperforms existing weakly supervised and most fully supervised methods, achieving high-precision segmentation that is critical for clinical diagnosis, treatment planning, and other workflows. 
\end{itemize}

\section{Related Work}
\subsection{Fully-supervised Learning Segmentation}
Fully supervised segmentation methods, particularly U-Net \cite{Unet, 3DUnet}, have excelled in medical image analysis due to their modular design and effectiveness in various modalities \cite{UNetsurvey2022}. 
Extensions of U-Net address scale and complexity in medical imaging \cite{Unet, 3DUnet, UNetsurvey2022}. U-Net++ \cite{UNet++} refines skip connections for multi-scale learning, DenseNet \cite{2016Densely} enhances feature representation, and V-Net \cite{VNet} accelerates 3D segmentation with residual blocks. DA-VNet \cite{meng20233d} applies dual-attention for cross-modal segmentation. Transformers capture long-range dependencies, with UNetR \cite{UNetR} integrating ViTs for global context while retaining CNN-based decoding. Structured state space models (SSMs) offer efficient long-range modeling \cite{mamba, mamba2}, and Vision Mamba (ViM) \cite{vim} improves performance with reduced memory usage.
However, most existing methods depend on labor-intensive pixel-wise annotations. Then, we turn to weakly supervised methods, which require only sparse annotations, which can significantly reduce annotation effort for clinicians and, thereby, enhance usability in clinical practice.

\subsection{Weakly-supervised Learning Segmentation}
Existing weakly-supervised learning segmentation methods can be roughly divided into three categories. The first category generated pseudo labels for unannotated pixels and is used to train the fully supervised segmentation model \cite{PASEG}. For example, ScribbleSup \cite{lin2016scribblesup} utilized a graphical model to propagate scribble annotations and optimized network parameters. 
WSS-CEC \cite{Dolz2022weakly} utilizes multi-modal images to generate enhanced class activation maps under image-level supervision. 
The method of supervision of mixed dynamic pseudo labels \cite{han2024dmsps} introduced a novel dual-branch network to dynamically generate pseudo labels for supervised training of segmentation models. 
The second category used regularization losses to constrain unannotated pixels and avoid overfitting, such as conditional random field loss \cite{PASEG}, entropy minimization loss \cite{EMLoss2004semi}, and total variation loss \cite{TVjavanmardi2016}, etc. 
The third category uses consistency learning strategies to ensure consistent predictions across different perturbations, thereby enhancing the robustness and reliability of segmentation results.
For instance, the transformation-consistent mean teacher model \cite{ustm2022PR} introduced a transformation-consistent learning technique based on a mean teacher framework to enhance the accuracy of segmentation. Scrbble2Label \cite{lee2020scribble2label} leveraged the consistency of the prediction results by label filtering to improve reliable pseudo labels from weak supervision. 

Most existing methods are designed for single modality and perform poorly on complex segmentation tasks, limiting their accuracy. Multi-modal approaches have the potential to address these challenges, but they remain underexplored.

\subsection{Cross-modal Fusion}
Cross-modal fusion is a key challenge in multi-modal learning, enabling the integration of features from different modalities into a common space. Most cross-modal medical image segmentation methods use input-level fusion, where multi-modal images are concatenated along the channel axis before training \cite{arrayzhou2019review}, as seen in MM-UNet \cite{Renarticle}.  
In hierarchical-level fusion, separate encoders process single or dual-modal inputs, and the extracted features are fused to enhance cross-modal information \cite{arrayzhou2019review}. HyperDense-Net \cite{dolz2018hyperdense} employs modality-specific branches with dense intra- and inter-path connections, while F²Net \cite{jbhiyang2023mmfusion} integrates shared representations for richer feature extraction.  
Decision-level fusion trains multiple segmentation networks separately and typically averages their predictions. MAML \cite{MAMLzhang2021modality} uses an ensemble of modality-specific models that learn collaboratively to handle unique features and missing modalities.

However, most existing methods rely on fully supervised tasks that require extensive pixel-wise annotations. In contrast, we aim to address modality fusion challenges in scribble-based weakly supervised learning to enhance segmentation performance. Additionally, we focus on capturing the unique characteristics of each modality and maximizing their complementary advantages through cross-modal fusion, further supporting weakly supervised learning.

\section{Method}
\subsection{Overview}
\begin{figure*}[!t]
\vspace{-0.1cm} 
\centerline{\includegraphics[trim={0cm 0cm 0cm 0.0cm},clip,width=15.5cm]{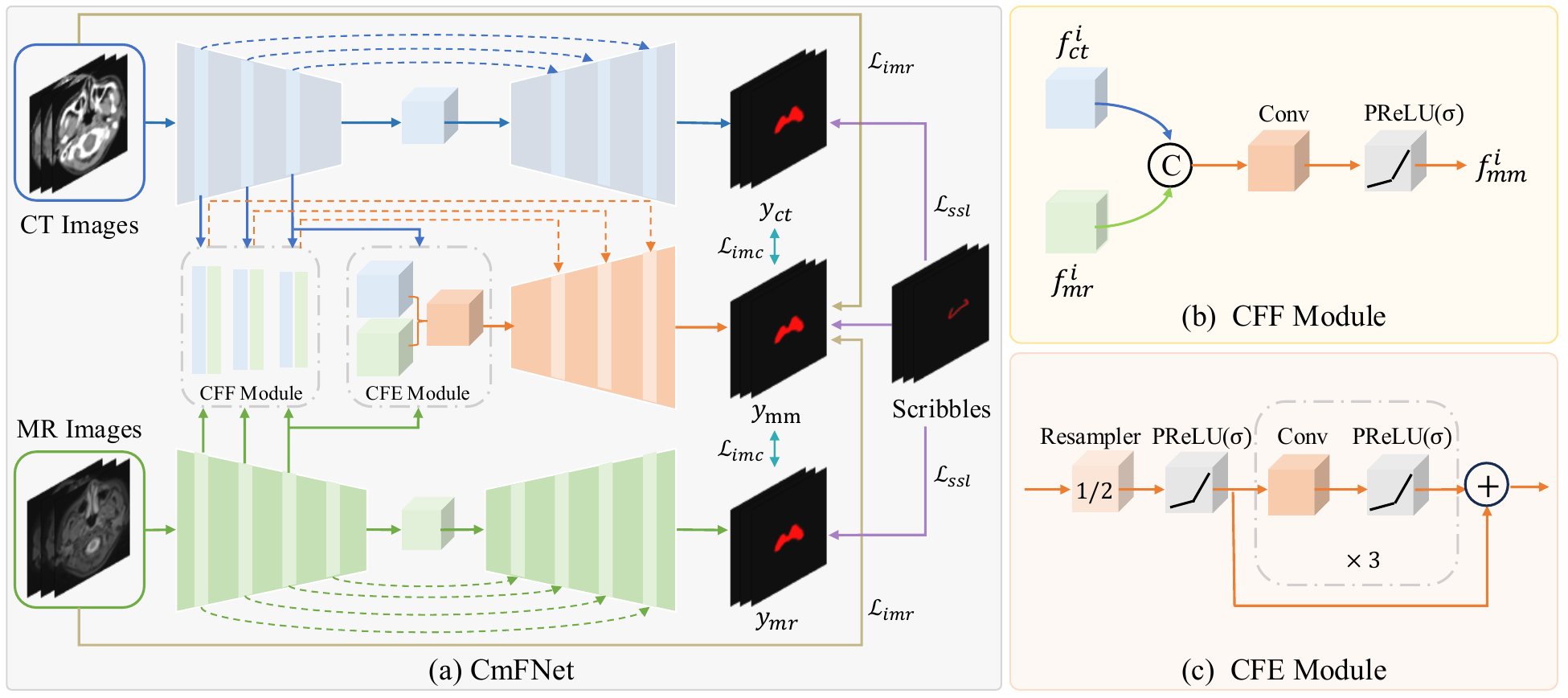}}
\caption{Overview of our approach. (a) Our network consists of triple branches, i.e., two modality-specific feature learning branches from individual modalities, along with a shared cross-modal feature learning branch. The prediction results of the three branches, $y_{ct}$, $y_{mr}$ and $y_{mm}$ are guided by a hybrid-supervised learning strategy that comprises scribble-supervised loss $L_{ssl}$, intra-modal regularization $L_{imr}$, and inter-modal consistency $L_{imc}$. The CFF (b) and CFE (c) are the cross-modal feature fusion module and the cross-modal feature enhancement module, respectively.}
\vspace{-0.1in}
\label{fig-network}
\end{figure*}

A schematic overview of the cross-modal fusion network (CmFNet) is shown in Fig.~\ref{fig-network}.
First, to prevent performance degradation caused by coarse scribble annotations, we adopt a cross-modal fusion strategy to incorporate finer details and enhance accuracy. We propose a triple-branch 3D network architecture consisting of two modality-specific feature learning branches and a shared cross-modal feature learning branch. To further refine segmentation, we introduce the Cross-Modal Feature Fusion (CFF) module, which integrates multi-scale features from modality-specific branches, and the Cross-Modal Feature Enhancement (CFE) module, which selectively enhances salient shared representations while leveraging complementary advantages from each modality. This design ensures optimal cross-modal information utilization, enabling precise target localization, capturing structural and edge details, and significantly improving segmentation performance. 

Secondly, to overcome the limitation of sparse scribble annotations for supervision, we introduce a hybrid-supervised learning strategy. 
This strategy integrates scribble supervision, intra-modal regularization, and inter-modal consistency constraints. 
Particularly, the intra-modal regularization is specifically designed to leverage the spatial relationships between pixels and contextual information, forcing the segmentation results of each modality to better conform to the internal relationships and distribution of the image, thereby mitigating the risk of overfitting.
Meanwhile, the inter-modal consistency aligns features from different modalities, ensuring coherence in feature representation and reducing modality discrepancies, enhancing robustness of cross modalities to ensure high-precision segmentation.

\subsection{Modality-specific Feature Learning}
Leveraging the unique advantages of diverse medical imaging modalities, we introduce a modality-specific feature learning network designed to effectively capture and represent modality-specific features.
Our network is built upon V-Net, a full convolutional volumetric neural network known for its end-to-end encoder-decoder design and 3D processing capabilities, which are particularly adept at capturing the spatial relationships and structures within 3D medical images \cite{VNet}. 

As depicted in Fig. \ref{fig-network} (a), the CT and MR volumes are independently fed into the two distinct branches. To optimize feature extraction for each modality, we customize down-sampling strategies. The CT branch, which undergoes three down-sampling operations, is tailored to retain detailed anatomical information essential for precise structural localization. In contrast, the MR branch, with four down-sampling stages, is designed to capture broader contextual information, enhancing the interpretation of soft tissue structures in relation to surrounding tissues.

These strategies yield two unique sets of multi-scale feature maps: $F_{ct}=[f_{ct}^{i}, i=1,...,3]$ for CT and $F_{mr}=[f_{mr}^{i}, i=1,...,4]$ for MR, where $i$ denotes the $i$th down-sampling operation, and after each down-sampling, the size of the feature map is reduced to half of its original dimensions. Significantly, these multi-scale feature maps are not only propagated to the decoder via skip connections to yield predictions for their respective branches but also furnish potent modality-specific feature information, facilitating the process of cross-modal feature learning. 

In the end, our modality-specific feature learning network leverages the unique capabilities of different imaging modalities, simultaneously driving effective cross-modal feature learning. This integrated strategy significantly boosts the performance and reliability of our segmentation method.

\subsection{Cross-modal Feature Learning}
As shown in Fig. \ref{fig-network} (a), the multi-scale modality-specific features derived from the two parallel single-modal branches are fed into the cross-modal feature learning branch. Subsequently, cross-modal shared feature fusion and enhancement are performed through the CFF module and CFE module, respectively, and then passed to the decoder path, culminating in the output of the predicted segmentation results.

\subsubsection{Cross-modal Feature Fusion (CFF) Module}
To maximize the advantages of modality-specific information, we use a plain CFF module to fuse the unique characteristics of each modality (see Fig.~\ref{fig-network} (b)). First, we define the unique feature maps from the two single-modal branches as $f_{ct} \in\mathbb{R}^{C \times H \times W \times D}$ and $f_{mr} \in\mathbb{R}^{C \times H \times W \times D}$, where $C$, $H$, $W$ and $D$ denote the channel, height, weight and depth, respectively. To validate the rich modality-specific features denoted by $f_{ct}$ and $f_{mr}$, we opt for a straightforward feature enhancement strategy instead of more complex approaches such as attention mechanisms \cite{meng20233d, jbhiyang2023mmfusion}. This design is crafted to showcase that the triple-branch segmentation network is not only proficient in effectively capturing modality-specific feature information but also in harmoniously fusing cross-modal feature information, thus eliminating the need for sophisticated architectural components. Specifically, two features of $i$th stage, $f_{ct}^{i}$ and $f_{mr}^{i}$ are first fused by a concatenation operation and then passed through a $3\times3\times3$ convolutional layer. The output of this layer is further passed through a PReLU activation function \cite{He2015Prelu}, as follows:
\begin{equation}
\begin{aligned}
f_{mm}^{i} = \sigma(Conv(Concat(f_{ct}^{i}, f_{mr}^{i}) \ ,
\end{aligned}
\end{equation}
where $i \in \{1, 2, 3\}$, $\sigma$ denotes the PReLU activation function, $Conv$ denotes convulutional operation and $Concat$ denotes a concatenation operation. Thus, the CFF module can effectively fuse cross-modal features under three down-sampling operations, thereby furnishing the subsequent network layers with rich multi-scale contextual information.

\subsubsection{Cross-modal Feature Enhancement (CFE) Module}
To alleviate the issue of feature misalignment arising from customized down-sampling strategies across modality-specific feature learning branches, we introduce a CFE module tailored to overcome this particular challenge. As shown in Fig.\ref{fig-network} (a), we select the maximum number of common down-sampling operations of the two modality-specific branches corresponding to the network layer, the third layer, and then pass the modality-specific feature maps of this layer to the CFE module accordingly. The architecture of the proposed CFE is shown in Fig.\ref{fig-network} (c). 
Moreover, to maximize the utility of cross-modal features and increase the receptive field, a down-sampling process is implemented on the enhanced feature maps.
\begin{equation}
\begin{aligned}
\widetilde{f}_{mm}^{3} = \sigma(DownConv(f_{mm}^{3})) \ ,
\end{aligned}
\end{equation}
where $f_{mm}^{3} = \sigma(Conv(Con 
cat(f_{ct}^{3}, f_{mr}^{3})$, $DownConv$ denotes convolution operations with a stride of 2. To further emphasize the important features, three convolutional modules are employed to process the cross-modal shared features. Thus, the CFE module not only effectively addresses the feature alignment issue, but also facilitates the cross-modal feature learning branch in capturing complementary cross-modal shared feature representations, thereby significantly bolstering segmentation performance.

\subsubsection{Decoder path}
The decoder within the cross-modal feature learning branch is designed to drive useful cross-modal shared features from the CFE module while integrating contextual information through the skip connections of the CFF module. It adeptly merges high-level semantic features with low-level detailed information, incrementally reconstructing the feature maps to the desired resolution and achieving accurate segmentation results.

\subsection{Hybrid-supervised learning}
By leveraging modality-specific and cross-modal learning branches, we can acquire rich and effective feature representations. However, relying solely on scribble annotations for supervision is insufficient, as they lack the detailed spatial and contextual information necessary for the precise segmentation of complex objects. 
Moreover, scribble supervision alone cannot ensure feature alignment across modalities, leading to inconsistencies that compromise model robustness and reliability.
To address these limitations, we propose a hybrid-supervised learning approach that combines scribble supervision, intra-modal regularization, and inter-modal consistency. This strategy enhances spatial and contextual learning, aligns cross-modal features, mitigates overfitting, and ensures stable, high-precision segmentation.

\subsubsection{Scribble-supervised Learning (SSL)}
We apply a partial cross-entropy (pCE) function for scribble-supervised learning, which is specifically designed for pixels with scribble annotation: 
\begin{equation}
\begin{aligned}
L_{pCE}(s, y) = - \sum_{c}\sum_{i \in \omega_{s}}\log y^{c}_{i} \ ,
\end{aligned}
\end{equation}
where $s$ is the one-hot scribble annotations, $ y^{c}_{i}$ is the predicted probability of pixel $i$ belonging class $c$, $\omega_{s}$ is the set of labeled pixels in $s$. Thus, our scribble-supervised loss is formulated as:
\begin{equation}
\begin{aligned}
\label{Eq-SLLloss}
L_{ssl}(s, y_{ct}, y_{mr}, y_{mm}) &= L_{pCE}(s, y_{ct}) + L_{pCE}(s, y_{mr}) \\
    & + L_{pCE}(s, y_{mm}) \ ,
\end{aligned}
\end{equation}
where $y_{ct}$ and $y_{mr}$ are the predictions of two modality-specific branches, respectively, and $y_{mm}$ is the prediction of the cross-modal branch. 

\subsubsection{Intra-modal regularization (IMR)}
By utilizing the inherent anatomical relationships and distributions within multi-modal images, we regularize the segmentation results within each modality, specifically highlighting the importance of spatial and contextual information among pixels. Firstly, we introduce the conditional random field (CRF)  \cite{PASEG}, which calculates the pairwise energy of each pixel pair to encourage consistent labels between pixels with similar features:
\begin{equation}
\begin{aligned}
L_{crf}(y) = \sum_{i,j \in \Omega}y_{i}(1-y_{j})K_{ij}  \ ,
\end{aligned}
\end{equation}
while $y_{i}$ and $y_{j}$ are the predictions of pixel $i$ and $j$, respectively, $\Omega$ is the set of pixels, and $K_{ij}=\sum^{N}_{n=1}\omega_{n}k_{n}(\textbf{f}^{n}_{i}, \textbf{f}^{n}_{j})$, which means the discontinuity cost based on a mixture on $N$ kernels, $\omega_{n}$ is the weight of the $n$-th kernel $k_{n}$ and $\textbf{f}^{n}$ is the feature vector of $k_{n}$.

Furthermore, to enable 3D contextual regularization, we use the multi-view CRF loss that applies the CRF in axial, sagittal, and coronal views, respectively:
\begin{equation}
\begin{aligned}
L_{mcrf}(y) = \frac{1}{3}(L^{A}_{crf}(y)+L^{S}_{crf}(y)+L^{C}_{crf}(y)) \ ,
\end{aligned}
\end{equation}
where $L^{A}_{crf}(y)$, $L^{S}_{crf}(y)$ and $L^{C}_{crf}(y)$ all follow $L^{'}_{crf} = \sum_{N}L_{crf}(y^{n})/N$, that is, the average 2D CRF loss across the D axial, W sagittal and H coronal slices, respectively. 

Then, the proposed intra-class regularization will constrain the prediction results through the images of each modality respectively:
\begin{equation}
\label{Eq-RLloss}
\begin{aligned}
L_{imr}(y_{mm}) = \lambda_{ct}L_{mcrf}(y_{mm}) +  \lambda_{ct}L_{mcrf}(y_{mm}) \ ,
\end{aligned}
\end{equation}
where $\lambda_{ct}$ and $\lambda_{mr}$ indicate the weights of the constraints from the CT images and MR images, respectively.

\subsubsection{Inter-modal consistency (IMC)}
To enhance feature alignment and mitigate modality discrepancies, we apply consistency constraints to the segmentation results across multiple modalities, thereby ensuring stable and robust results. Specifically, we calculate the Mean Squared Error (MSE) between the predictions of the cross-modal branch and the single-modal branch, promoting collaborative learning:
\begin{equation}
\begin{aligned}
\label{Eq-CLloss}
L_{imc} = \alpha_{1}\Vert y_{mm}-y_{ct}\Vert^{2} + \alpha_{2}\Vert y_{mm}-y_{mr}\Vert^{2}  \ ,
\end{aligned}
\end{equation}
where $\alpha_1$ and $\alpha_2$ represent the constraint weights between modalities. Specifically, the multi-modal branch can achieve better segmentation results than the single-modal branch under ideal conditions, making it a suitable reference value.

Finally, the overall hybrid-supervised learning strategy for training the cross-modal segmentation model is summarized as:
\begin{equation}
\begin{aligned}
\label{Eq-Totalloss}
L_{total} = L_{ssl} + L_{imr} + L_{imc} \ .
\end{aligned}
\end{equation}
Therefore, this strategy effectively uses intra-modal regularization to constrain the relationship between pixels, preventing overfitting. Moreover, by ensuring inter-modal consistency, it yields robust and highly precise segmentation results.

\section{Experiments}
\subsection{Datasets}
We evaluated our method through two types of tasks: 1) segmentation of nasopharyngeal carcinoma tumors within a clinical cross-modal dataset denoted as NPC, and 2) segmentation of whole abdominal organs-at-risk from a public CT dataset denoted as WORD \cite{luo2022word}. 
For comparisons of weakly supervised segmentation methods, we used only scribble annotations for the training dataset. Examples of two datasets and their corresponding annotations are shown in Fig. \ref{fig-samples}.

\subsubsection{NPC Dataset}
The NPC dataset contains 161 patients who received radiotherapy treatment at a Cancer Hospital. This study was performed under the institutional review board (IRB) approval
of Beijing Cancer Hospital (IRB No. 2020KT99).
The CT images were reconstructed using a matrix size of $512\times512$, thickness of 3.0 $mm$, and pixel size of 1.27 $\times$ 1.27 $mm^{2}$. The MR T2 images were reconstructed using a matrix size of $384\times384$, thickness of 3.0 $mm$, and pixel size of 1.30 $\times$ 1.30 $mm^{2}$. The manual segmentation of NPC was contoured by a radiation oncologist and verified by an experienced oncologist. In our study, we utilized a five-fold cross-validation strategy to divide and facilitate model training on the NPC dataset.

\subsubsection{WORD Dataset}
The WORD dataset includes 150 CT scans from patients who received radiotherapy treatment at a single center \cite{luo2022word, miccai2023word}. The CT images were reconstructed as $512\times512$, thickness of 2.5 $mm$ to 3.0 $mm$, and pixel size of 0.976 $\times$ 0.976 $mm^{2}$. In our study, we followed the default settings in \cite{luo2022word}, splitting the WORD dataset into three parts: 100 scans for training, 20 scans for validation, and 30 scans for testing. We tested our method on 7 organs: the liver, spleen, left kidney, right kidney, stomach, gallbladder and pancreas. 

\subsubsection{Scribble Generation}
For the NPC dataset, we used ITK-SNAP \cite{itksnap} to manually draw scribbles on CT images for both the tumor and background within the segmentation mask, following the principles \cite{valvano2021TMI} to guarantee realistic scribble annotations. Furthermore, to reduce the burden of manual annotation, we also adopted an eroding method to automatically generate the background scribble annotations and conducted comparative experiments to validate its effectiveness. The average image coverages of scribbles for the background, auto-generated background, and tumor target were $0.08\%$, $0.06\%$, and $0.04\%$, respectively. For the WORD dataset, annotations with scribbles of abdominal organs were provided by \cite{luo2022word}.

\subsubsection{Evaluation Metric}
To quantitatively evaluate the segmentation performance, we utilized the Dice Similarity Coefficient (DSC) and Average Surface Distance (ASD) as evaluation metrics to compare the prediction results with the ground truth.

\subsection{Implementation Details}
The model was implemented with the PyTorch \cite{paszke2019pytorch} framework on one NVIDIA A6000 GPU.
For our ablation study, all models were trained from scratch with the same experimental settings. The training used the Adam optimizer with a learning rate of $1\times10^{-5}$, batch size of 16, 200 epochs, and dropout of 0.5. The $\lambda_{1}$, $\lambda_{2}$, $\alpha_1$, and $\alpha_2$ were set to 0.2, 0.2, 0.8, and 0.8, respectively.
For a fair comparison, we used publicly available source codes and followed the original experimental settings to train all models from scratch. During inference, the output from the cross-modal feature learning branch served as the final segmentation result.
In the training procedure, the input images were randomly cropped as a 3D volume with size $96\times96\times16$. In the inference procedure, the input images with arbitrary shapes were fed into the trained model directly.
 
\subsection{Comparison with SOTA Methods}
To comprehensively evaluate our method, we compared it with various SOTA including both weakly supervised and fully supervised segmentation methods. Notably, throughout the following tables present, the best-performing method is highlighted in bold, while the second-best is marked in blue.

\subsubsection{Ours Vs. Weakly-supervised Methods on NPC Dataset}
We first compared our method to several SOTA scribble-supervised segmentation methods in the cross-modal imaging scenario, including 1) baseline V-Net \cite{VNet} with only partial cross-entropy loss (pCE) \cite{lin2016scribblesup}, 2) Pseudo-label generation strategy, such as dynamically mixed soft pseudo-labels supervision (DMSPS) \cite{han2024dmsps}, 3) Different regularization strategies, including entropy minimization (EM) \cite{EMLoss2004semi}, total variation (TV) \cite{TVjavanmardi2016} and mumford-shah (MS) \cite{MSLoss}, 4) Different consistency learning strategies, including transformation-consistent mean teacher model (USTM) \cite{ustm2022PR} and scribble2Label(S2L) \cite{lee2020scribble2label}.
In addition, we employ fully supervised annotations based on the cross-entropy loss, to establish an upper bound performance (FullySup).
Most weakly supervised segmentation methods are designed for single-modal images. To ensure a fair comparison, we merged multiple modalities along the channel dimension before feeding them to the segmentation models, following the strategy in \cite{jbhiyang2023mmfusion}.

\begin{figure*}[!t]
\vspace{-0.1cm} 
\centerline{\includegraphics[trim={0cm 0cm 0cm 0.2cm},clip,width=16.0cm]{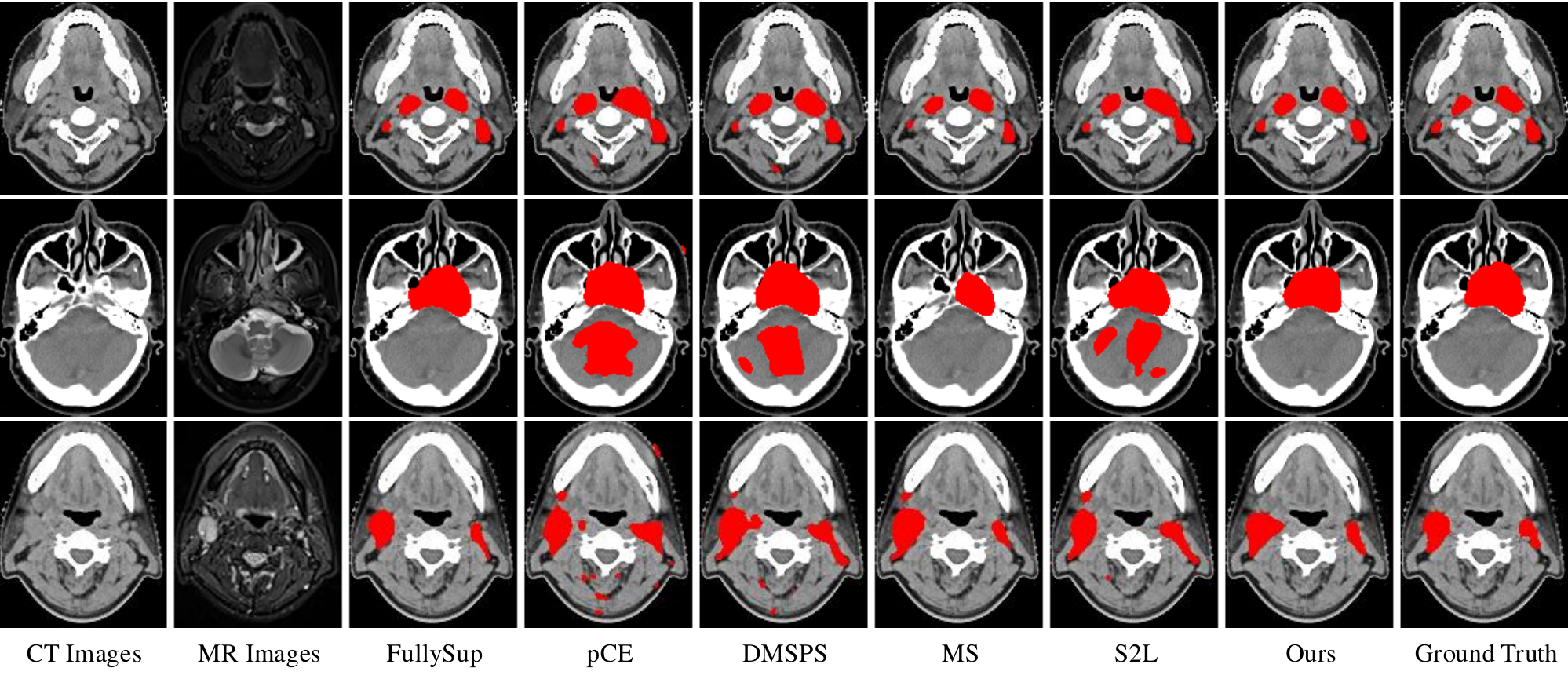}}
\caption{Qualitative comparison between our method and SOTA weakly-supervised methods on the NPC dataset (corresponding to Table \ref{Table-wss-npc}). Our approach achieves the best performance and is very close to the ground truth.}
\vspace{-0.1in}
\label{fig-wss-npc}
\end{figure*}

Table \ref{Table-wss-npc} and Fig. \ref{fig-wss-npc} show the qualitative and quantitative evaluation of weakly-supervised segmentation results. As indicated in Table \ref{Table-wss-npc}, our method outperforms the SOTA methods, achieving high-accuracy segmentation results. In addition, when using automatically generated background annotations instead of manual annotations (AutoBG), our method still achieves consistent performance. Fig. \ref{fig-wss-npc} shows that our method effectively identifies tumor locations, with results closely matching the actual tumor regions. Meanwhile, other methods may struggle to handle complex structural information, incorrectly identifying normal tissues as tumors, resulting in significant discrepancies from the ground truth.
Therefore, our method successfully models the spatial and contextual information of anatomical structures, and is able to fully utilize consistent cross-modal feature representations, ultimately achieving reliable and high-accuracy segmentation results.

\begin{table}
\caption{Quantitative comparison between our method and SOTA weakly-supervised methods on NPC dataset. }
\label{Table-wss-npc} 
\setlength{\tabcolsep}{9pt}
\begin{tabular}{l|l|lll}
\hline\noalign{\smallskip}
    Methods  & Data & DSC ($\%$) $\uparrow$  & ASD  {(mm)} $\downarrow$ \\
     \noalign{\smallskip}\hline\noalign{\smallskip}
    FullySup & masks & 73.98 $\pm$ 7.28  & 3.71 $\pm$ 1.62 \\
    pCE \cite{lin2016scribblesup} & scribbles & 57.38 $\pm$ 13.12& 13.78 $\pm$ 5.72  \\
    DMSPS \cite{han2024dmsps} & scribbles & 71.58 $\pm$ 7.73 & 4.70 $\pm$ 1.81 \\
    EM \cite{EMLoss2004semi} & scribbles & 69.91 $\pm$ 8.79 & 5.23 $\pm$ 2.25  \\
    TV \cite{TVjavanmardi2016} & scribbles & 74.55 $\pm$ 7.30 & 3.25 $\pm$ 1.31 \\
    MS \cite{MSLoss} & scribbles & 73.91 $\pm$ 7.21 & 3.53 $\pm$ 1.41 \\
    USTM \cite{ustm2022PR} & scribbles & 69.04 $\pm$ 7.99& 5.55 $\pm$ 1.98  \\
    S2L \cite{lee2020scribble2label} & scribbles & 70.97 $\pm$7.57& 6.11 $\pm$ 3.31 \\
    Ours (AutoBG) & scribbles & \color{blue}75.39 $\pm$ 5.90 & \color{blue}2.95 $\pm$ 0.95\\ 
    Ours & scribbles & \pmb{76.75 $\pm$ 5.78} & \pmb{2.69 $\pm$ 0.89} \\    
\noalign{\smallskip}\hline
\end{tabular}
\vspace{-0.1in}
\end{table}

\subsubsection{Ours Vs. Weakly-supervised Methods on WORD Dataset}
WORD dataset only contains CT images.
In handling this situation, we input CT images into both the MR and CT branches while keeping all other experimental settings unchanged.

\begin{figure*}[!t]
\vspace{-0.1cm} 
\centerline{\includegraphics[trim={0cm 0cm 0cm 0.2cm},clip,width=16.0cm]{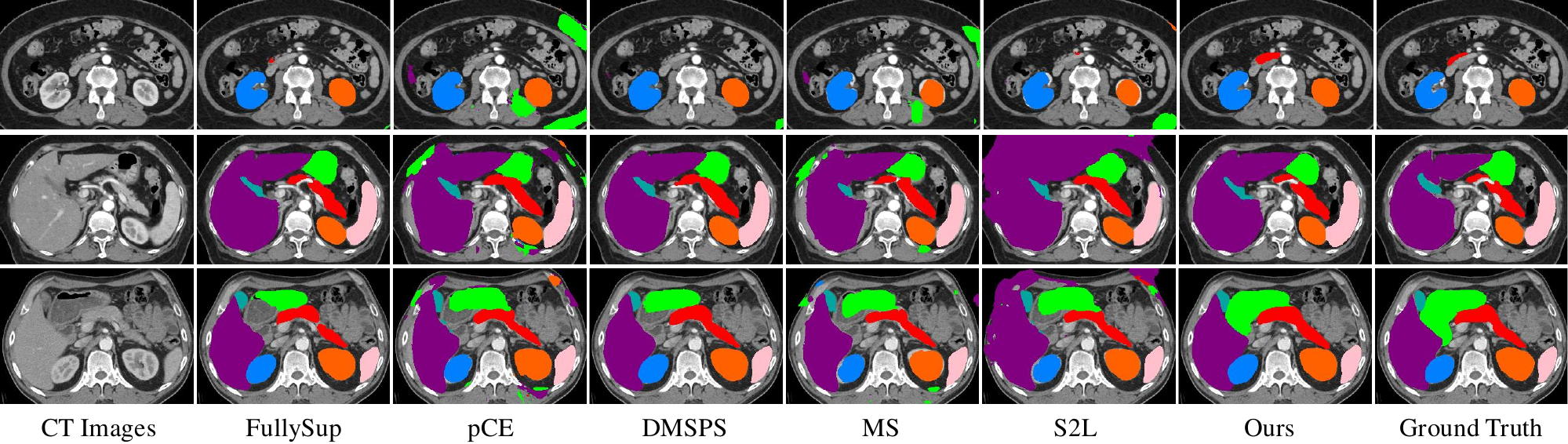}}
\caption{Qualitative comparison between our method and SOTA weakly-supervised methods on the WORD dataset (corresponding to Table \ref{Table-wss-word}).}
\vspace{-0.1in}
\label{fig-wss-word}
\end{figure*}

Table \ref{Table-wss-word} and Fig. \ref{fig-wss-word} present the results obtained on the WORD dataset. Specifically, our approach generally ranks the second best, only behind the task-specific DMSPS \cite{han2024dmsps}. Notably, while our method is not specific for single-modal scenarios, it still delivers exceptional segmentation performance in terms of ASD metrics for organs such as the kidney, gallbladder, and pancreas, even surpassing the task-specific DMSPS \cite{han2024dmsps}. Moreover, our method achieved strong performance for the stomach and pancreas, which have complex spatial and shape variations, outperforming most weakly supervised methods in Fig. \ref{fig-wss-word}. This demonstrates that our method can effectively capture the complex variations of anatomical structures in 3D space and utilize shared feature information between branches to achieve robust segmentation. Overall, our approach is applicable to both cross-modal and single-modal data.

\begin{table*}
\caption{Quantitative comparison between our method and SOTA weakly-supervised methods on WORD dataset. our approach generally ranks the second best, only behind the task-specific DMSPS \cite{han2024dmsps}.}
\centering
\label{Table-wss-word} 
\begin{adjustbox}{width=\linewidth, keepaspectratio}
\begin{tabular}{l|l|l|lllllll|l}
\hline\noalign{\smallskip}
      Metric & Methods  & Data & Liver & Spleen & Kidney (L) & Kidney(R) & Stomach & Gallbladder & Pancreas & Mean \\
     \noalign{\smallskip}\hline\noalign{\smallskip}
    \multirow{8}{*}{DSC ($\%$) $\uparrow$ } & FullySup & masks & 96.41 $\pm$ 0.81 & 95.40 $\pm$ 1.61 & 94.92 $\pm$ 1.79 & 95.19 $\pm$ 1.92 & 89.51 $\pm$ 4.68 & 77.87 $\pm$ 11.36 & 81.46 $\pm$ 7.49 & 90.11 $\pm$ 4.24\\
    & pCE \cite{lin2016scribblesup} & scribbles & 82.03 $\pm$ 3.54 & 88.05 $\pm$ 4.71 & 86.47 $\pm$ 5.25 & 83.11 $\pm$ 5.31 & 66.02 $\pm$ 11.52 & 61.37 $\pm$ 17.91 & 60.35 $\pm$ 9.01 & 75.34 $\pm$ 8.18 \\
    & DMSPS \cite{han2024dmsps} & scribbles & \pmb{94.90 $\pm$ 0.71} & \pmb{92.96 $\pm$ 2.38} & \pmb{92.71 $\pm$ 1.91} & \pmb{92.91 $\pm$ 1.87} & \pmb{87.40 $\pm$ 3.93} & \pmb{69.19 $\pm$ 14.36} & \pmb{77.03 $\pm$ 7.03} & \pmb{86.72 $\pm$ 4.60 } \\
    & EM \cite{EMLoss2004semi} & scribbles & 87.60 $\pm$ 2.90 & 88.83 $\pm$ 3.93 & 88.39 $\pm$ 3.58 & 90.67 $\pm$ 2.09 & 71.62 $\pm$ 10.71 & 65.59 $\pm$ 16.80 & 72.00 $\pm$ 7.55 & 80.67 $\pm$ 6.79 \\
    & TV \cite{TVjavanmardi2016} & scribbles & 84.88 $\pm$ 3.67 & 84.72 $\pm$ 6.58 & 86.04 $\pm$ 4.67 & 84.27 $\pm$ 4.68 & 66.62 $\pm$ 10.41 & 49.35 $\pm$ 20.11 & 67.99 $\pm$ 8.91 & 74.84 $\pm$ 8.43  \\
    & MS \cite{MSLoss} & scribbles & 84.81 $\pm$ 2.81 & 86.95 $\pm$ 4.36 & 86.39 $\pm$ 4.08 & 89.34 $\pm$ 2.19 & 71.64 $\pm$ 8.68 & 59.59 $\pm$ 20.28 & 70.72 $\pm$ 7.98 & 78.49 $\pm$ 7.20  \\
    & USTM \cite{ustm2022PR} & scribbles & 85.83 $\pm$ 4.62 & 89.82 $\pm$ 2.99 & 90.28 $\pm$ 2.88 & 91.48 $\pm$ 2.29 & 80.19 $\pm$ 5.64 & 63.72 $\pm$ 18.16 & \color{blue}75.11 $\pm$ 6.85 & 82.35 $\pm$ 6.20 \\
    & S2L \cite{lee2020scribble2label} & scribbles & 71.56 $\pm$ 4.04 & \color{blue}90.80 $\pm$ 3.13 & 86.92 $\pm$ 4.12 & 88.29 $\pm$ 5.86 & 78.87 $\pm$ 5.81 & 59.82 $\pm$ 20.24 & 69.60 $\pm$ 8.64 & 77.98 $\pm$ 7.41 \\
    & Ours & scribbles & \color{blue}90.93 $\pm$ 5.92 & 88.45 $\pm$ 4.91 &  \color{blue}91.98 $\pm$ 2.55 &  \color{blue}92.38 $\pm$ 2.27 &  \color{blue}85.01 $\pm$ 7.01 &  \color{blue}67.32 $\pm$ 19.14 & 73.29 $\pm$ 9.61 &  \color{blue}84.19 $\pm$ 7.34 \\
    \noalign{\smallskip}\hline
    \multirow{8}{*}{ASD {(mm)} $\downarrow$} & FullySup & masks & 1.01 $\pm$ 0.90 & 1.60 $\pm$ 3.52 & 0.72 $\pm$ 0.33 & 0.71 $\pm$ 0.29 & 2.34 $\pm$ 1.80 & 3.36 $\pm$ 4.57 & 1.88 $\pm$ 1.84 & 1.66 $\pm$ 1.89 \\
    & pCE \cite{lin2016scribblesup} & scribbles & 27.56 $\pm$ 10.76 & 10.27 $\pm$ 7.18 & 5.59 $\pm$ 4.48 & 42.03 $\pm$ 10.65 & 49.00 $\pm$ 17.80 & 16.14 $\pm$ 9.86 & 37.97 $\pm$ 7.26 & 26.94$\pm$ 9.71 \\
    & DMSPS \cite{han2024dmsps} & scribbles & \pmb{1.62 $\pm$ 1.13} & \pmb{1.52 $\pm$ 1.26} & \pmb{1.10 $\pm$ 0.65} & {1.75 $\pm$ 1.74} & \pmb{3.30 $\pm$ 3.73} & \color{blue}{4.89 $\pm$ 5.24} & \color{blue}{2.80 $\pm$ 2.11} & \pmb{2.43 $\pm$ 2.27} \\
    & EM \cite{EMLoss2004semi} & scribbles & 12.04 $\pm$ 8.24 & 5.83 $\pm$ 4.48 & 5.27 $\pm$ 4.92 & 2.14 $\pm$ 1.41 & 39.36 $\pm$ 20.84 & 12.61 $\pm$ 13.13 & 13.79 $\pm$ 7.90 & 13.01 $\pm$ 8.70 \\
    & TV \cite{TVjavanmardi2016} & scribbles & 17.10 $\pm$ 8.88 & 22.04 $\pm$ 10.77 & 12.73 $\pm$ 10.75 & 43.10 $\pm$ 13.04 & 49.03 $\pm$ 18.47 & 44.39 $\pm$ 15.30 & 23.16 $\pm$ 7.31 & 30.22 $\pm$ 12.07 \\
    & MS \cite{MSLoss} & scribbles & 12.13 $\pm$ 10.12 & 12.70 $\pm$ 11.63 & \color{blue}4.79 $\pm$ 6.72 & 3.42 $\pm$ 3.31 & 26.24 $\pm$ 16.92 & 19.53 $\pm$ 14.81 & 10.02 $\pm$ 7.53 & 12.69 $\pm$ 10.15 \\
    & USTM \cite{ustm2022PR} & scribbles & 16.28 $\pm$ 7.39 & 3.83 $\pm$ 4.02 & 5.12 $\pm$ 7.51 & 3.77 $\pm$ 5.79 & 13.79 $\pm$ 10.56 & 20.70 $\pm$ 42.81 & 6.31 $\pm$ 3.76 & 9.91 $\pm$ 11.69 \\
    & S2L \cite{lee2020scribble2label} & scribbles & 17.66 $\pm$ 8.09 & 3.05 $\pm$ 3.45 & 8.43 $\pm$ 8.50 & \color{blue}1.68 $\pm$ 1.61 & 15.52 $\pm$ 14.14 & 16.97 $\pm$ 25.24 & 7.43 $\pm$ 5.02 & 10.11 $\pm$ 9.44 \\
    & Ours & scribbles &\color{blue} 2.02 $\pm$ 1.28 & \color{blue}2.15 $\pm$ 2.81 & 4.93 $\pm$ 6.32 & \pmb{1.44 $\pm$ 2.91} & \color{blue}3.42 $\pm$ 2.90 & \pmb{4.09 $\pm$ 4.82} & \pmb{2.41 $\pm$ 1.63} & \color{blue}2.92 $\pm$ 3.15 \\
    \noalign{\smallskip}\hline
\end{tabular}
\vspace{-0.1in}
\end{adjustbox}
\end{table*}

\subsubsection{Ours Vs. Fully-supervised Methods}
We also compared our method with several SOTA fully-supervised segmentation methods on the NPC dataset, including 1) specialized cross-modal segmentation methods for head and neck tumors: M-CNN \cite{Ma_2019}, MM-UNet \cite{Renarticle} and DANet \cite{meng20233d}, 2) the most common SOTA segmentation methods, including V-Net \cite{VNet}, ResUNet \cite{ResUnet}, UNet++ \cite{UNet++}, DenseNet \cite{2016Densely} and UNetR \cite{UNetR}. Notably, our approach, along with all the compared methods, utilizes full annotations. For a comprehensive evaluation, we referred to the experimental results from \cite{meng20233d} on the NPC dataset. 

\begin{figure*}[!t]
\vspace{-0.1cm} 
\centerline{\includegraphics[trim={0cm 0cm 0cm 0.2cm},clip,width=16.0cm]{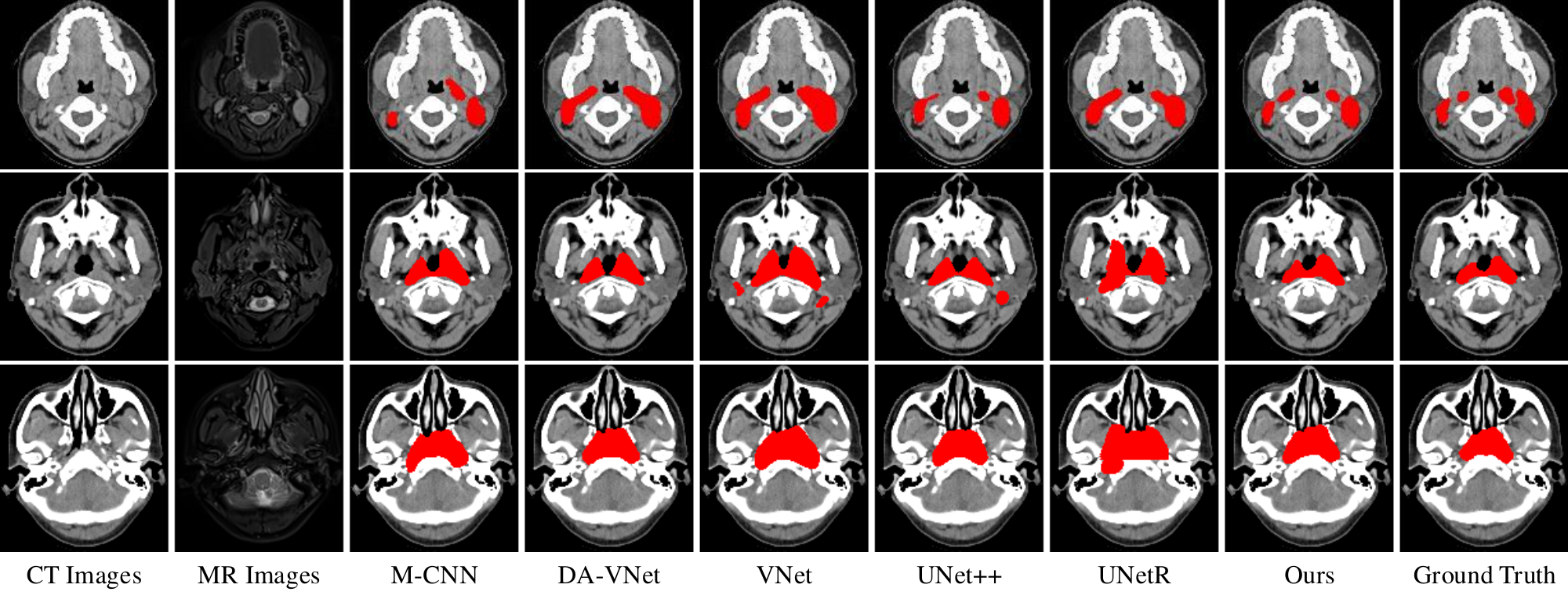}}
\caption{Qualitative comparison between our method and SOTA fully-supervised methods on NPC dataset (corresponding to Table \ref{Table-fss-npc}).}
\vspace{-0.1in}
\label{fig-fss}
\end{figure*}

Table \ref{Table-fss-npc} presents a qualitative comparison of fully supervised methods, where our approach outperforms others in DSC and ASD metrics. Fig. \ref{fig-fss} visualizes the results, showing our method closely aligning with the ground truth. Notably, our method excels at accurately identifying and segmenting dispersed tumors, outperforming other approaches that fail to effectively capture multi-modal features, resulting in detail loss and imprecise segmentation. By leveraging modality-specific and cross-modal feature learning branches, our method also demonstrates robustness in fully supervised segmentation, highlighting its potential in medical image analysis.

\begin{table}
\centering
\caption{Quantitative comparison between our method and SOTA fully supervised methods on NPC dataset.}
\label{Table-fss-npc} 
\setlength{\tabcolsep}{9pt}
\begin{tabular}{l|l|lll}
\hline\noalign{\smallskip}
    Methods  & Data & DSC ($\%$) $\uparrow$  & ASD  {(mm)} $\downarrow$ \\
     \noalign{\smallskip}\hline\noalign{\smallskip}
    M-CNN \cite{Ma_2019} & masks & 73.72 $\pm$ 8.51 & 3.11 $\pm$ 1.35 \\
    MM-UNet \cite{Renarticle} & masks  & 72.49 $\pm$ 7.32 & 2.99 $\pm$ 11.09 \\
    DA-VNet \cite{meng20233d} & masks  & \color{blue}77.58 $\pm$ 6.75 & \color{blue}2.52 $\pm$ 0.98 \\
    V-Net \cite{VNet}  & masks & 73.98 $\pm$ 7.28  & 3.71 $\pm$ 1.62 \\
    ResUNet  \cite{ResUnet} & masks & 73.53 $\pm$ 9.77 & 3.06 $\pm$ 1.43 \\
    U-Net++ \cite{UNet++}  & masks & 73.01 $\pm$ 11.71 & 3.25 $\pm$ 2.04 \\
    DenseNet \cite{2016Densely}  & masks & 71.95 $\pm$ 2.41 & 4.17 $\pm$ 2.41 \\
    UNetR \cite{UNetR} & masks  & 67.40 $\pm$ 12.10 & 3.72 $\pm$ 1.62 \\
    Ours & masks  & \pmb{78.95 $\pm$ 5.99} & \pmb{2.33 $\pm$ 0.80} \\
\noalign{\smallskip}\hline
\end{tabular}
\vspace{-0.1in}
\end{table}

\subsection{Ablation Study}
We conduct ablation studies on the modality-specific feature learning, cross-modal feature learning, hybrid-supervised learning strategy, and hyper-parameters setting in our network, respectively.

\subsubsection{Modality-specific Feature Learning}
We first assessed the performance of single-modal CT and MR models under different network architectures, including variations in layer depth and down-sampling (Ds) operations.
Subsequently, we analyzed the segmentation performance of the modality-specific feature learning network in different combined strategies, using the SSL and IMC as learning objectives.

Table \ref{Table-single-depth} summarizes the performance of single-modal models with varying network architectures. The CT$_{3}$ model, with fewer down-sampling steps, preserves anatomical details, achieving the highest DSC score. However, increasing down-sampling from three to four reduces DSC performance. Meanwhile, the deeper CT$_{6}$ model achieves superior ASD, highlighting the importance of deeper architectures for capturing complex features and improving tumor boundary precision. Similar trends are observed in single-modal MR models.

\begin{table}
\centering
\caption{Quantitative comparison of single-modal segmentation models under different network architectures.}
\label{Table-single-depth} 
\setlength{\tabcolsep}{4pt}
\begin{tabular}{l|l|l|l|ll}
\hline\noalign{\smallskip}
     Methods & Depth & Ds Count & Params & DSC ($\%$) $\uparrow$ & ASD  {(mm)} $\downarrow$ \\
     \noalign{\smallskip}\hline\noalign{\smallskip}
    CT$_3$ & conv3 & 3 & 7.94M  & \pmb{74.35 $\pm$ 6.09} & 3.39 $\pm$ 1.33 \\
    CT$_4$ & conv4 & 4 & 19.57M & 73.66 $\pm$ 6.18 & 3.40 $\pm$ 1.17 \\
    CT$_5$ & conv5 & 4 & 35.57M & 73.16 $\pm$ 6.60 & 3.44 $\pm$ 1.22 \\
    CT$_6$ & conv6 & 4 & 51.56M & 73.02 $\pm$ 7.85 & \pmb{3.34 $\pm$ 1.78} \\
    \noalign{\smallskip}\hline
    MR$_3$ & conv3 & 3 & 7.94M  & \pmb{70.86 $\pm$ 8.79} & 3.93 $\pm$ 1.43 \\
    MR$_4$ & conv4 & 4 & 19.57M & 70.34 $\pm$ 9.33 & 3.81 $\pm$ 1.38 \\
    MR$_5$ & conv5 & 4 & 35.57M & 70.10 $\pm$ 10.06 & 3.79 $\pm$ 1.59 \\
    MR$_6$ & conv6 & 4 & 51.56M & 70.58 $\pm$ 10.96 & \pmb{3.45 $\pm$ 1.40} \\
\noalign{\smallskip}\hline
\end{tabular}
\vspace{-0.1in}
\end{table}

Table \ref{Table-multimodal-depth} details the performance of modality-specific feature learning models with different combined strategies. It reveals that the strategy of combining CT$_{6}$MR$_4$ and CT$_{3}$MR$_4$ yields the optimal performance for $y_{ct}$ and $y_{mr}$, respectively. 
Furthermore, CT$_{3}$MR$_4$ achieves a more balanced segmentation across $y_{ct}$ and $y_{mr}$.
By minimizing down-sampling operations in the CT branch, CT$_{3}$MR$_4$ preserves detailed anatomical information, thereby enhancing the modeling of spatial and contextual relationships. Meanwhile, increasing depth and down-sampling in the MR branch enables the learning of sophisticated feature representations, refining target edge segmentation. This collaborative learning between modality-specific branches leads to mutual performance gains. In contrast, CT$_{6}$MR$_4$ configuration shows a more pronounced performance difference in  $y_{ct}$ and $y_{mr}$. This discrepancy is due to the deeper CT and MR branches learning similar features and losing detailed information. Ultimately, CT$_{3}$MR$_4$ excels by learning complementary modality-specific representations, thus enhancing cross-modal learning.

\begin{table*}
\centering
\caption{Quantitative comparison of
modality-specific feature learning segmentation models with different combined strategies. CT$_{6}$MR$_4$ and CT$_{3}$MR$_4$ achieved the best performance for $y_{ct}$ and $y_{mr}$, respectively.}
\label{Table-multimodal-depth} 
\begin{tabular}{l|l|llll|llll}
\hline\noalign{\smallskip}
      \multirow{2}{*}{Branches} & \multirow{2}{*}{Methods} & \multicolumn{4}{c}{DSC ($\%$) $\uparrow$} & \multicolumn{4}{c}{ASD {(mm)} $\downarrow$} \\
     & & CT$_3$ & CT$_4$ & CT$_5$ & CT$_6$ & CT$_3$ & CT$_4$ & CT$_5$ & CT$_6$\\
    \noalign{\smallskip}\hline\noalign{\smallskip}
     \multirow{4}{*}{\makecell[c]{$y_{ct}$}} & MR$_3$ & 73.72 $\pm$ 6.41 & 73.99 $\pm$ 6.25 & 73.37 $\pm$ 7.04 & 73.42 $\pm$ 7.57 & 4.04 $\pm$ 1.78 & 3.46 $\pm$ 1.42 & 3.50 $\pm$ 1.54 & 3.33 $\pm$ 1.38 \\
     & MR$_4$ & 73.76 $\pm$ 6.64 & 73.05 $\pm$ 6.43 & 74.13 $\pm$ 6.65 & \pmb{74.51 $\pm$ 6.60} & 3.87 $\pm$ 1.64 & 3.58 $\pm$ 1.38 & 3.25 $\pm$ 1.24 & \pmb{3.23 $\pm$ 1.27} \\
     & MR$_5$ & 73.04 $\pm$ 7.11 & 73.00 $\pm$ 6.73 & 73.37 $\pm$ 7.04 & 73.58 $\pm$ 7.48 & 4.52 $\pm$ 1.91 & 3.80 $\pm$ 1.66 & 3.50 $\pm$ 1.54 & 3.37 $\pm$ 1.47  \\
     & MR$_6$ & 73.18 $\pm$ 6.51 & 72.81 $\pm$ 6.73 & 73.77 $\pm$ 6.82 & 74.46 $\pm$ 6.42 & 4.42 $\pm$ 2.11 & 3.78 $\pm$ 1.61 & 3.31 $\pm$ 1.42 & 3.29 $\pm$ 1.23 \\
     \noalign{\smallskip}\hline
     \multirow{4}{*}{\makecell[c]{$y_{mr}$}} & MR$_3$ & 70.79 $\pm$ 9.13 & 66.83 $\pm$ 8.79 & 70.30 $\pm$ 11.06 & 66.86 $\pm$ 8.54 & 3.84 $\pm$ 1.53 & 6.76 $\pm$ 2.85 & 3.72 $\pm$ 1.54 & 6.31 $\pm$ 2.55 \\
     & MR$_4$ & \pmb{71.14 $\pm$ 9.42} & 70.04 $\pm$ 10.66 & 69.40 $\pm$ 8.65 & 70.08 $\pm$ 8.69 & \pmb{3.59 $\pm$ 1.51} & 3.72 $\pm$ 1.89 & 4.75 $\pm$ 1.73 & 4.13 $\pm$ 1.49 \\
     & MR$_5$ & 71.08 $\pm$ 10.02 & 70.29 $\pm$ 8.77 & 70.00 $\pm$ 9.22 & 69.77 $\pm$ 9.20 & 3.60 $\pm$ 1.41 & 4.23 $\pm$ 1.46 & 4.37 $\pm$ 1.70 & 4.29 $\pm$ 1.50\\
     & MR$_6$ & 70.65 $\pm$ 9.87 & 70.43 $\pm$ 9.75 & 70.04 $\pm$ 10.38 & 70.22 $\pm$ 9.13 & 3.79 $\pm$ 1.62 & 4.50 $\pm$ 2.41 & 3.93 $\pm$ 1.60 & 4.40 $\pm$ 1.78\\
\noalign{\smallskip}\hline
\end{tabular}
\vspace{-0.1in}
\end{table*}

\subsubsection{Cross-modal Feature Learning}
Three models are used to evaluate the cross-modal feature learning network: the baseline modality-specific feature learning (MFL) network, one with CFE and a specialized decoder, and another with CFF for fusing modality-specific features with multi-scale context.

Table \ref{Table-multimodal-fusion} presents the quantitative evaluation results, demonstrating the effectiveness of cross-modal feature learning. The CFE module and decoder leverage complementary cross-modal features, significantly improving segmentation performance. Meanwhile, the CFF module enriches multi-scale contextual information at different decoder stages, achieving optimal results. By incorporating cross-modal learning, our approach preserves modality-specific features while maximizing cross-modal information. The CFF and CFE modules further enhance feature fusion, boosting segmentation accuracy.

\begin{table}
  \centering
\caption{Quantitative evaluation of modules in cross-modal feature learning, confirming the effectiveness of each module.}
\label{Table-multimodal-fusion} 
\setlength{\tabcolsep}{4pt}
\begin{tabular}{l|lll|ll}
\hline\noalign{\smallskip}
     Models & MFL & CFE & CFF & DSC ($\%$) $\uparrow$  & ASD  {(mm)} $\downarrow$  \\
     \noalign{\smallskip}\hline\noalign{\smallskip}
        MFL & $\checkmark$ & $\times$ & $\times$ & 73.76 $\pm$ 6.64 & 3.87 $\pm$1.64 \\
        MFL + CFE & $\checkmark$ & $\checkmark$ & $\times$ & 75.96 $\pm$ 5.94 & 2.80 $\pm$ 0.89 \\
        MFL + CFE + CFF & $\checkmark$ & $\checkmark$ & $\checkmark$ &  \pmb{76.75 $\pm$ 5.78} & \pmb{2.69 $\pm$ 0.89} \\
\noalign{\smallskip}\hline
\end{tabular}
\vspace{-0.1in}
\end{table}

\subsubsection{Hybrid-supervised Learning}
We evaluated the effectiveness of two key learning objects in our method: the IMR strategy and the IMC strategy. 
As shown in Table \ref{Table-loss}, compared to using only the SSL strategy, the IMR regularizes the segmentation results within each modality and mitigates overfitting by concentrating on the spatial and contextual relationships between pixels, thereby improving the ability to locate and segment complex targets.
In addition, the IMC achieves stable and high-precision segmentation results by promoting feature alignment between modalities through consistency constraints.
Ultimately, our approach integrates the IMR and IMC strategies, guiding the learning process through intra-modal regularization and inter-modal consistent constraints. This hybrid-supervised strategy not only mitigates overfitting but also achieves superior segmentation performance.

\begin{table}
\caption{Quantitative evaluation of hybrid-supervised learning strategies used in our approach.}
\label{Table-loss} 
\setlength{\tabcolsep}{4pt}
\begin{tabular}{l|lll|ll}
\hline\noalign{\smallskip}
    Models & SSL & IMR & IMC & DSC ($\%$) $\uparrow$  & ASD  {(mm)} $\downarrow$ \\
     \noalign{\smallskip}\hline\noalign{\smallskip}
    SSL & $\checkmark$ & $\times$ & $\times$ & 68.20 $\pm$ 8.56 & 8.04 $\pm$ 3.85  \\
    SSL + IMR  & $\checkmark$ & $\checkmark$ & $\times$ & 75.79 $\pm$ 6.43 & 2.86 $\pm$ 1.17  \\
    SSL + IMC & $\checkmark$ & $\times$ & $\checkmark$ & 76.45 $\pm$ 5.96 & 2.81 $\pm$ 0.97  \\
    SSL + IMR + IMC & $\checkmark$ & $\checkmark$ & $\checkmark$ & \pmb{76.75 $\pm$ 5.78} & \pmb{2.69 $\pm$ 0.89} \\
\noalign{\smallskip}\hline
\end{tabular}
\vspace{-0.1in}
\end{table}

\subsubsection{Hyper-parameters}
\begin{figure}[!t]
\vspace{-0.1cm}
  \centering
    \subfigure
    {
        \includegraphics[width=\columnwidth]{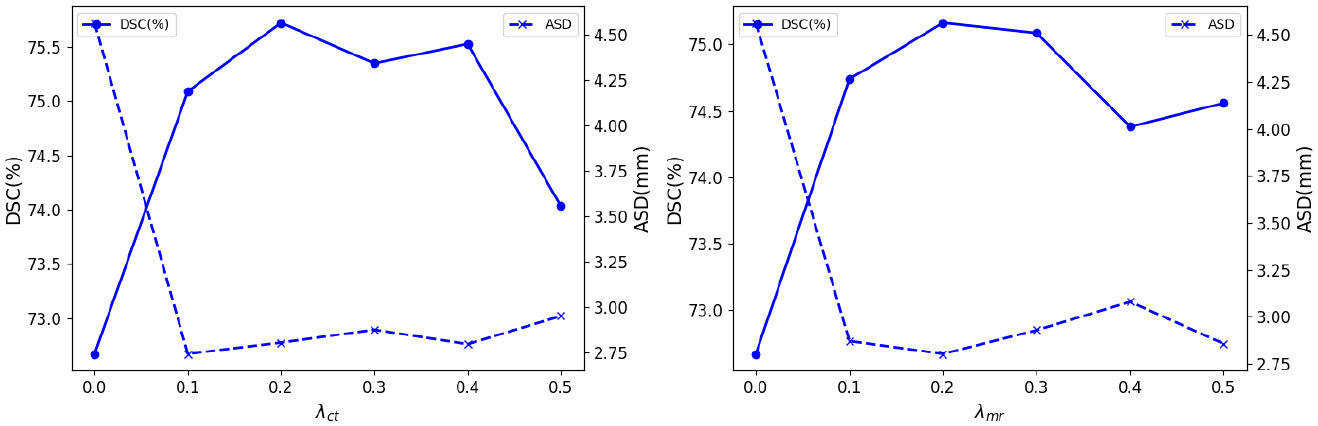}
    }
    \subfigure
    {
         \includegraphics[width=\columnwidth]{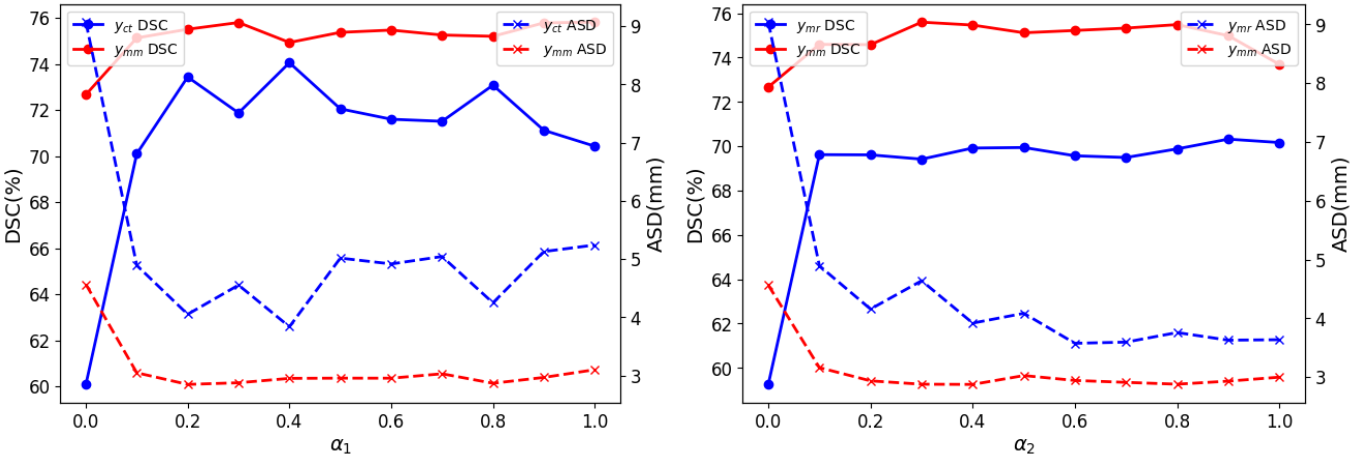}
    }
\caption{Sensitivity analysis of hyper-parameter $\lambda_{ct}$, $\lambda_{mr}$, $\alpha_1$ and $\alpha_2$ on the NPC dataset, respectively.}
\vspace{-0.2in}
\label{fig-hyperpara}
\end{figure}

Our method is characterized by four hyper-parameters: $\lambda_{ct}$ and $\lambda_{mr}$ as specified in Eq.(\ref{Eq-RLloss}), which govern the influence of the IMR, and $\alpha_1$ along with $\alpha_2$ in Eq.(\ref{Eq-CLloss}), which regulate the weights assigned to the IMC. We evaluated the performance by adjusting $\lambda_{ct}$ and $\lambda_{mr}$ across a spectrum from 0.0 to 0.5. Fig. \ref{fig-hyperpara} shows the trend of the average scores $DSC$ and $ASD$ throughout the NPC dataset. It should be noted that assigning a value of 0.2 to both $\lambda_{ct}$ and $\lambda_{mr}$ produces a well-balanced performance in the evaluation metrics $DSC$ and $ASD$.

We also evaluated our method’s performance with $\alpha_1$ and $\alpha_2$ set within (0.0, 1.0). As shown in Fig. \ref{fig-hyperpara}, increasing $\alpha_1$ and $\alpha_2$ improves both the cross-modal and modality-specific branches, though the gain stabilizes over time. Notably, the cross-modal branch consistently outperforms the modality-specific branch due to its effective integration of complementary cross-modal information. Setting $\alpha_1$ and $\alpha_2$ to 0.8 achieves an optimal balance between the branches.

\section{Conclusion and Discussion}
In this paper, we present a novel method for 3D weakly-supervised cross-modal medical image segmentation that reduces annotation costs while maintaining high-precision segmentation performance. 
Our method effectively addresses the challenges of accuracy degradation and overfitting in weakly supervised learning, particularly in cases involving small tumor regions and complex anatomical structures.
The modality-specific feature learning network effectively enhances segment accuracy by capturing essential characteristics from various modalities, benefiting both cross-modal and single-modal data.
Furthermore, our cross-modal feature learning effectively uses complementary information from two modalities, resulting in minimal degradation in accuracy for our weakly-supervised method compared to the fully-supervised method.
Additionally, our hybrid-supervised learning strategy successfully eliminates the risk of overfitting under weak supervision by modeling the spatial contextual relationships and ensuring feature alignment, which boosts consistency and stable segmentation results. This can facilitate clinical therapy and benefit various specialists, including physicists, radiologists, pathologists, and oncologists. 

Our efforts also offer insights into maximizing mutual information gained by using specific network architectures and leveraging modality complementarity through structured information bridges to facilitate both intra- and inter-feature transfer and enhancement. These designs make complex segmentation tasks accessible, and similar improvements can be anticipated when applied to other challenging tasks in multi-modal medical imaging beyond CT and MRI.
In future work, we will develop adaptive strategies to identify key features of each modality in optimal network architectures, streamline model complexity, improve modality fusion, and advance practical clinical applications.

\bibliographystyle{ieeetr}
\bibliography{ref}
\end{document}